%% file: main.tex
\documentclass[letterpaper, 10 pt, conference]{ieeeconf}
\IEEEoverridecommandlockouts

\input{preamble.tex}

\begin{document}

\title{Bayesian Continuum Robot Dynamics and State Estimation}

\author{
James~M.~Ferguson,$^{1}$~Tucker~Hermans,$^{2}$~and~Alan~Kuntz$^{1}$%
\thanks{
Research reported in this publication was supported by the Advanced Research Projects Agency for Health (ARPA-H) under the ALISS project, Award Number D24AC00415-00. The ARPA-H award of up to \$11,935,038 provided 100\% of the financial support for this work.
The opinions and findings in this paper are solely the responsibility of the authors and do not necessarily represent the official views of ARPA-H.
}%
\thanks{
$^{1}$Department of Electrical and Computer Engineering and Department of Computer Science, Vanderbilt University, TN, USA.
$^{1}$Kahlert School of Computing and Robotics Center, University of Utah, UT, USA.
{\tt\footnotesize james.m.ferguson@vanderbilt.edu}}}

\maketitle

\begin{abstract}

Recent factor graph approaches to continuum robot state estimation have been successful for quasi-static applications and spatiotemporal estimation using white-noise kinematic motion priors.
However, when inertial effects are significant, these approximations may fail to capture the underlying physics, limiting accuracy during dynamic motions.
In contrast, our approach approximates the Cosserat rod dynamics of continuum robots.
We write inertia and damping as equivalent applied loads, so that the dynamic balance retains the algebraic form of the static one from prior work with quasi-static robots.
Without backbone observations, the framework reduces to a stochastic forward simulation of the robot's motion.
Given observations, it jointly refines kinematic and dynamic states and infers external loads, among other states.
We validate the approach through simulation and experiments, demonstrating stochastic forward simulation as well as state estimation on tendon-driven continuum robots.

\end{abstract}

\maketitle

\section{Introduction}
\label{sec:introduction}

Continuum robots are a class of compliant manipulators whose backbones deform continuously in response to actuation and external loading. 
Their slender geometry enables dexterous operation in confined environments, making them ideal candidates for surgical applications \cite{burgner2015continuum, russo2023continuum}. 
A typical modeling objective is to estimate equilibrium state at a single moment in time (e.g., shape, external loads) given noisy observations (e.g., actuation, tip pose) \cite{anderson2017continuum, lilge2022continuum, ferguson2024unified, lilge2024state, lilge2025incorporating, ferguson2026continuum, prakash2026multi}.
Such quasi-static assumptions are justified when robot dynamics are relatively fast and damped or of low amplitude relative to the task at hand.  

However, many continuum robots do exhibit significant dynamics responses, motivating dynamic models, rather than quasistatic \cite{gravagne2003large, till2019dynamics, rone2013continuum, rucker2011statics, renda2018discrete, boyer2021dynamics}.
Knowledge of dynamics enables more accurate tracking and control when the effects of inertia are significant (e.g. larger or more compliant manipulators) \cite{fischer2023dynamic, aner2023modeling, mishra2023trajectory, yang2024differential, song2015shape}.
These approaches are typically solved using constant curvature models \cite{webster2010design} or Partial Differential Equation (PDE) solver approaches \cite{till2019dynamics, alessi2024rod} that do not take uncertainty or online backbone observations into account.

\begin{figure}
    \centering
    \vspace{0pt}
    \hspace{-20pt}
    \input{figures/front_page.tikz}
    \vspace{-5pt}
    \caption{Spatiotemporal continuum robot dynamics state estimation with continuum robot dynamics. \textit{Top}: a continuum robot undergoing dynamic motion, with states evolving over space and time. \textit{Bottom}: the fixed-lag smoother factor graph. Each time step holds a full quasi-static robot graph \cite{ferguson2026continuum} with Cosserat spatial factors (red) alongside actuation, external load, and measurement factors (Fig.~\ref{fig:robot_graph}); time steps are coupled by temporal kinematics, compatibility, and dynamics factors (yellow). States leaving the window are marginalized into a history prior (green).}
    \label{fig:intro_figure}
    \vspace{-10pt}
\end{figure}
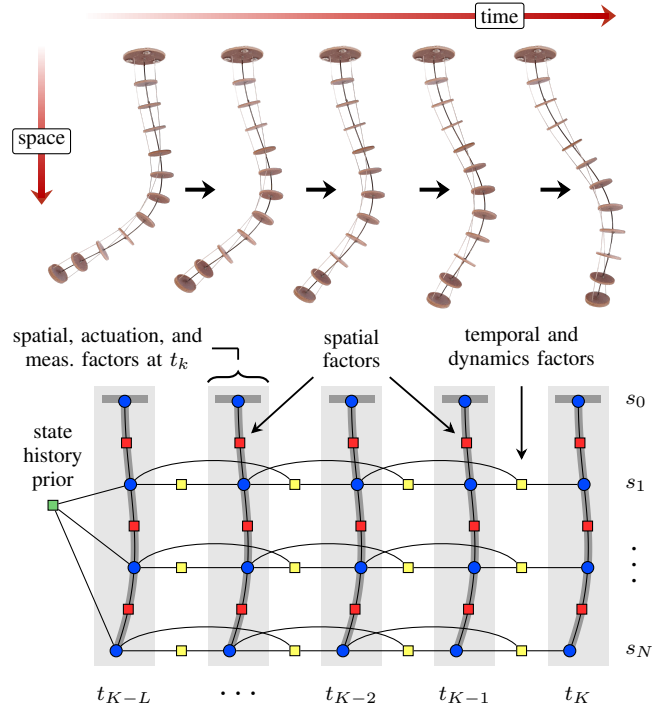
\begin{figure*}
    \centering
    \input{figures/robot_graph.tikz}
    \vspace{-20pt}
    \caption{Factor graph for a single time step of robot motion, only showing 5 discs for simplicity; steps are strung together by the temporal graph in Fig.~\ref{fig:intro_figure}. Given priors on tensions $\brm{q}_k$, base pose $\pose_{1,k}$, and tip load $\wrench_k^\mathrm{e}$, the model solves a discrete form of the robot dynamics including inertia and damping. Optional measurement factors (e.g., tip position) refine the estimate, including the tip load.}
    \label{fig:robot_graph}
    \vspace{-15pt}
\end{figure*}
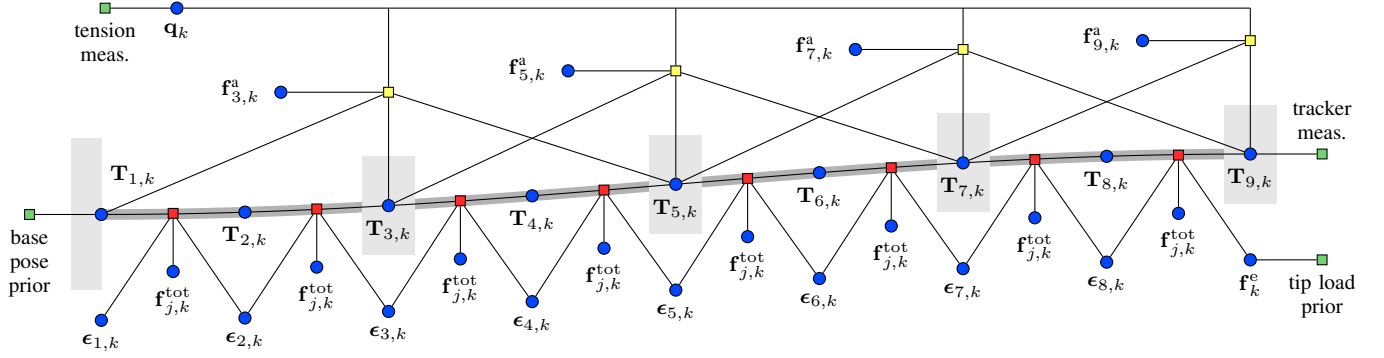

In the context of Bayesian state estimation, Teetaert et al. introduced
spatiotemporal kinematic motion priors for continuum
robots~\cite{teetaert2025stochastic, teetaert2026sliding}.
Their white-noise-on-acceleration priors are an elegant and efficient formulation
that yields accurate estimates given sufficient measurements, and they partially
address the challenges of spatiotemporal estimation.
However, being kinematic, these priors do not capture actuation or inertial
effects, which matter when measurements are sparse or the motion is dynamic.

To address this, we extend the quasi-static factor graph framework
of~\cite{ferguson2026continuum} to dynamics, replacing kinematic priors with
factors derived from Cosserat rod dynamics that model actuation, inertia, and
damping.
The graph is solved as a \textit{fixed-lag smoother}: optimization runs over a
sliding time window and older states are marginalized out, giving a joint
distribution over shape, velocity, strain, actuation, and external loads within
the window.
Posing the dynamics as inference, rather than as a PDE solve with measurements
attached afterward, has three benefits.
First, the solver reports uncertainty at every timestep, indicating not only the
estimated shape but how far it can be trusted.
Second, each sensor enters as a factor on whichever variables it observes, so
modalities as different as a tip tracker and a shape-sensing fiber can be fused
without changing the dynamics model.
Third, marginalization bounds the cost per step while retaining the information
from all past measurements.

In this paper, we present a factor graph formulation of Cosserat rod dynamics, with actuation,
inertia, and damping, solved as a fixed-lag smoother.
We show that, without measurements, it reproduces a benchmark dynamic
solver~\cite{till2019dynamics} using far fewer spatial nodes, while also providing
uncertainty.
We show in simulation that, given noisy tip position measurements, it estimates
tip forces accurately and consistently where a quasi-static estimator does not.
Finally, on an open-source dynamic-motion dataset~\cite{gotelli2026multi}, we show
that it estimates the robot's shape from actuation alone and that adding a fiber
Bragg grating shape sensor~\cite{shi2016shape} further improves the estimate.

Overall, our approach provides a single probabilistic framework for continuum
robots undergoing dynamic motion: without measurements, it acts as an accurate
forward simulator with uncertainty, and with measurements, it performs principled
Bayesian state estimation.

\section{Cosserat Rod Dynamics}
\label{sec:dynamics}

We build directly on the quasi-static discrete Cosserat rod model from prior work~\cite{ferguson2026continuum}, retaining its conventions.
Poses $\pose=\{\rot,\pos\}\in\SE$ map body to spatial coordinates, and twists $\bm{\xi}=(\bm{\omega},\bm{\nu})\in\se$ are stored with their rotational component first, inducing the adjoints
\begin{equation}
    \Ad(\pose)
    =
    \begin{bmatrix}
        \rot & 0 \\
        \hatop{\pos}\rot & \rot
    \end{bmatrix},
    \qquad
    \curly{\bm{\xi}}
    =
    \begin{bmatrix}
        \hatop{\bm{\omega}} & 0 \\
        \hatop{\bm{\nu}} & \hatop{\bm{\omega}}
    \end{bmatrix}.
\label{eq:adjoints}
\end{equation}
The generalized strain deviates from its stress-free value $\strain_0$
through the linear constitutive law
\begin{equation}
\stress=\begin{bmatrix}\brm{m}\\ \brm{n}\end{bmatrix}
       =\stiff\,(\strain - \strain_0),
\label{eq:constitutive}
\end{equation}
relating the generalized stress $\stress$---internal moment $\brm{m}$ and force $\brm{n}$---to strain through the rod stiffness $\stiff\in\mathbb{R}^{6\times6}$ encoding sensitivity to bending, torsion,
shear, and elongation.

\subsection{Spatiotemporal Kinematics}

We build directly on the spatiotemporal work of Teetaert et. al \cite{teetaert2025stochastic}.
During dynamic motion, the backbone pose evolves both spatially with arclength $s$ and temporally with time $t$; i.e., $\pose(s,t) \in \SE$.
Throughout this paper, we omit the function arguments, with the understanding that all states implicitly depend on $s$ and $t$.

Following \cite{teetaert2025stochastic, teetaert2026sliding}, we introduce the body-frame temporal velocity $\vel\in\se$.
Combined with the strain $\strain$, the spatiotemporal kinematics are
\begin{equation}
\frac{\partial\pose}{\partial s}=\pose \, \hatop{\strain},
\qquad
\frac{\partial\pose}{\partial t}=\pose \, \hatop{\vel}.
\label{eq:kinematics}
\end{equation}
Whereas the strain $\strain$ can be interpreted as a velocity with respect to arclength (e.g., \cite{ferguson2026continuum, lilge2022continuum}), $\vel$ is the standard temporal velocity associated with robot motion \cite{teetaert2025stochastic}.

Because $\strain$ and $\vel$ are both derivatives of the same pose field, they cannot be chosen independently.
Equating the mixed partials and reducing yields the compatibility condition \cite{rucker2011statics, teetaert2025stochastic}
\begin{equation}
\frac{\partial\vel}{\partial s}
=\frac{\partial\strain}{\partial t}-\curly{\strain}\vel .
\label{eq:compatibility}
\end{equation}
This condition is an integrability requirement.
Since we estimate $\pose$, $\strain$, and $\vel$ as separate variables, \eqref{eq:compatibility} must be enforced explicitly \cite{teetaert2025stochastic}, as in Section~\ref{sec:compat_factor}.


\subsection{Momentum Balance}

For a Cosserat rod undergoing dynamic motion, the general equations of motion take the form \cite{simo1988dynamics, till2019dynamics, teetaert2025stochastic}
\begin{equation}
\begin{aligned}
    \stiff \frac{\partial \strain}{\partial s}
    &= 
    \curlyT{\strain} \stiff (\strain - \strain_0)
    - \wrench
    \\
    &+ \tfrac{\partial}{\partial t}(\inertia\vel)
    - \curlyT{\vel}(\inertia\vel)
    + \damping\,\vel
    .
\label{eq:dynamic_balance}
\end{aligned}
\end{equation}
The first line of \eqref{eq:dynamic_balance} is essentially that of \cite{lilge2022continuum, ferguson2026continuum}, transporting $\strain$ and adding the wrench $\wrench$, which is now distributed over both space and time.
The second line adds both inertia and damping; $\inertia$ is the generalized inertia matrix, and $\damping$ is linear viscous damping.

\subsection{Wrench Decomposition}

Moving the inertial and damping terms to the load side
returns~\eqref{eq:dynamic_balance} to the form of the static
balance,
\begin{equation}
\begin{aligned}
    \stiff \frac{\partial \strain}{\partial s} &=
    \curlyT{\strain} \stiff (\strain - \strain_0) - \wrenchtot,
    \\[0.5em]
    \wrenchtot &= 
    \wrenchext+\wrenchact+\wrenchgrav+\wrenchin .
    \label{eq:wrench_sum}
\end{aligned}
\end{equation}
Now the total distributed wrench collects four contributions: external loads $\wrenchext$, tendon actuation
$\wrenchact$ (Section~\ref{sec:actuation}), gravity $\wrenchgrav$, and the
inertial wrench (Section~\ref{sec:inertial}), which we define
\begin{equation}
\wrenchin=\curlyT{\vel}(\inertia\vel)
        -\frac{\partial}{\partial t}(\inertia\vel)
        -\damping\,\vel .
\label{eq:inertial_wrench}
\end{equation}
The first two terms are the D'Alembert inertial load \cite{lanczos1970variational}. 
Damping is a genuine dissipative load rather than a fictitious one, but is likewise velocity-dependent, so we account for it in the same equation and later in our factors.
We assume that gravity $\wrenchgrav$ is a known, deterministic load, entering $\wrenchtot$ directly so that it remains separate from the estimated external wrench $\wrenchext$.

\section{Discrete Estimation Approach}
\label{sec:discrete}

Our approach represents the robot's motion as a single factor graph that spans
both space and time. At each timestep, spatial factors connect states at discrete
points along the robot's length (Fig.~\ref{fig:robot_graph}), and temporal factors
connect these states across consecutive timesteps (Fig.~\ref{fig:intro_figure}).
This section defines each factor and describes how the overall factor graph is assembled and optimized for spatiotemporal continuum robot dynamics estimation.

We discretize the backbone into $s_j \in \{1 \dots N_s \}$ arclength nodes with uniform spacing $\Delta s$, and the trajectory into $t_k \in \{1 \dots N_t \}$ time steps, also with uniform spacing $\Delta t$.
Each node carries a pose $\pose_{j,k}\in\SE$, a strain $\strain_{j,k}\in\se$, a velocity $\vel_{j,k}\in\se$, and wrench variables.

\subsection{Piecewise-Linear Magnus Integration}
\label{sec:magnus}

Both kinematic equations in~\eqref{eq:kinematics} share the same form: a pose driven along one axis of $(s,t)$ by a body-frame rate:
\begin{equation}
\frac{\partial\brm{X}}{\partial\tau}=\brm{X}\,\hatop{\bm{\zeta}(\tau)},
\qquad \tau\in[0,h].
\label{eq:generic_ode}
\end{equation}
A common approximation holds $\bm{\zeta}$ constant over the interval, but as noted in \cite{ferguson2026continuum, barfoot2024state} a piecewise-linear model provides significantly higher numerical accuracy. 
Given endpoint values
$\bm{\zeta}_a,\bm{\zeta}_b$ we interpolate with
\begin{equation}
    \bm{\zeta}(\tau)=\bm{\zeta}_a+\tau\Delta\bm{\zeta}/h
    ,\quad
    \Delta\bm{\zeta}=\bm{\zeta}_b-\bm{\zeta}_a. 
\end{equation}
The solution to the group integration can be approximated by the highly accurate Magnus expansion \cite{blanes2009magnus}.
Truncating at fourth order \cite{barfoot2024state} gives the \emph{effective rate}
\begin{equation}
\magnus(\bm{\zeta}_a,\bm{\zeta}_b,h)
  =\bm{\zeta}_a+\tfrac{1}{2}\Delta\bm{\zeta}
  -\tfrac{h}{12}\,\curly{\Delta\bm{\zeta}}\bm{\zeta}_a
  +\tfrac{h^{2}}{240}\big(\curly{\Delta\bm{\zeta}}\big)^{2}\bm{\zeta}_a ,
\label{eq:magnus}
\end{equation}
so that $\brm{X}(h) \approx \brm{X}(0)\exp(\hatop{\magnus}h)$.
Retaining only the first term recovers the usual piecewise-constant integration; the remaining terms correct for the fact that the rates at the two endpoints do not commute.
Since the same expansion serves both axes of $(s,t)$, one implementation is used for the spatial and temporal kinematics factors.

\subsection{Spatial Kinematics Factor}
\label{sec:spatial_kin}

\begin{figure}
    \centering
    \vspace{0pt}
    \includegraphics[width=1.0\linewidth, clip, trim=30cm 20cm 18cm 20cm]{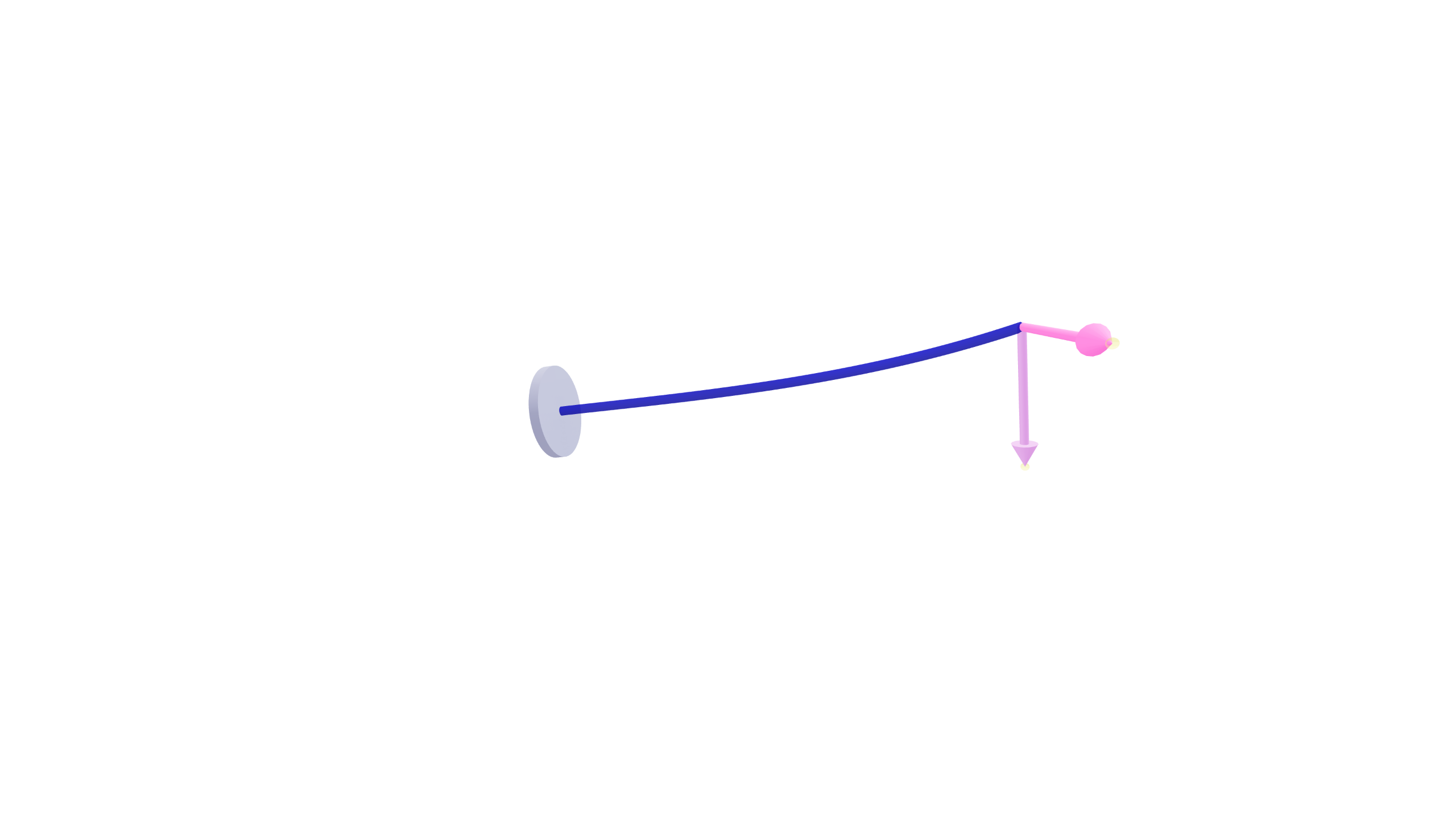}
    \includegraphics[width=0.9\linewidth, clip, trim=0cm 5cm 0cm 0cm]{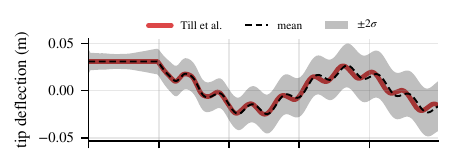}
    \includegraphics[width=0.9\linewidth, clip, trim=0cm 0cm 0cm 7.1cm]{figures/cosserat_rod_example_results.pdf}
    \\[0.5em]
    \includegraphics[width=0.9\linewidth, clip, trim=0cm 0cm 0cm 2.6cm]{figures/cosserat_rod_example_results.pdf}
    \vspace{-15pt}
    \caption{Stochastic forward simulation of a Cosserat rod using our method. A tip wrench (violet force, pink moment) is held for $0.1$~s and released, and the rod rings freely. \textit{Top}: Tip deflection along the swinging axis, against Till et al. \cite{till2019dynamics}. \textit{Bottom}: mean and standard deviation of that deflection over space and time, plotted as a heatmap. RMS tip position error between our solver and \cite{till2019dynamics} was \textbf{2.80 mm}.}
    \label{fig:cosserat_rod_example}
    \vspace{-10pt}
\end{figure}

%
%

Applying Section~\ref{sec:magnus} to the first equation of~\eqref{eq:kinematics} over the spatial interval $[s_j,s_{j+1}]$ at a fixed time $t_k$ gives
\begin{equation}
\pose_{j+1,k}=\pose_{j,k}\exp\!\Big(\big[
   \magnus(\strain_{j,k},\strain_{j+1,k},\Delta s)+\brm{n}_{\strain}
   \big]^{\wedge}\Delta s\Big),
\label{eq:spatial_prop}
\end{equation}
with $\brm{n}_{\strain}\sim\mathcal{N}(0,\bm{\Sigma}_{\strain})$.
That is, the pose at the next node is advanced along the rod by the effective rate that the two endpoint strains imply.
Solving for the noise yields the residual
\begin{equation}
\res^{\strain}_{j,k}
  =\magnus\big(\strain_{j,k},\,\strain_{j+1,k},\,\Delta s\big)
  -\frac{1}{\Delta s}\ln\big(\pose_{j,k}^{-1}\pose_{j+1,k}\big)^{\vee},
\label{eq:spatial_kin_factor}
\end{equation}
with $J^{\strain}_{j,k}=\|\res^{\strain}_{j,k}\|^{2}_{\bm{\Sigma}_{\strain}^{-1}}$.
The residual is therefore the disagreement, per unit arclength, between the strain the rod is estimated to carry and the strain implied by the relative pose of its neighboring nodes.
$\bm{\Sigma}_{\strain}$ accounts for departures of the shape, per unit arclength\new{,} from the constitutive law (e.g. from inexact $\stiff$)\new{.}
This is the kinematics factor of \cite{ferguson2026continuum}.

Note that we write this and every factor below in terms of strain; in code we implement them using stress $\stress$, equivalent by~\eqref{eq:constitutive} but better conditioned in the stiff shear and elongation directions.
All residual Jacobians follow from the sum, product, and chain rules applied to GTSAM's Lie group derivatives~\cite{dellaert2012factor}.

\subsection{Temporal Kinematics Factor}

The velocity plays the same role in time that strain plays in arclength, so the second equation of~\eqref{eq:kinematics} discretizes identically over the temporal interval $[t_k,t_{k+1}]$ at fixed $s_j$:
\begin{equation}
\res^{\vel}_{j,k}
  =\magnus\big(\vel_{j,k},\,\vel_{j,k+1},\,\Delta t\big)
  -\frac{1}{\Delta t}\ln\big(\pose_{j,k}^{-1}\pose_{j,k+1}\big)^{\vee},
\label{eq:temporal_kin_factor}
\end{equation}
with $J^{\vel}_{j,k}=\|\res^{\vel}_{j,k}\|^{2}_{\bm{\Sigma}_{\vel}^{-1}}$, reusing the same implementation.
$\bm{\Sigma}_{\vel}$ accounts for motion over $\Delta t$ that the node's velocity does not explain.
In our experiments, we set $\bm{\Sigma}_{\vel} = \bm{\Sigma}_{\strain}$ for simplicity.

\subsection{Compatibility Factor}
\label{sec:compat_factor}

The compatibility condition~\eqref{eq:compatibility} follows from the
mixed-partial identity $\partial_t\partial_s\pose=\partial_s\partial_t\pose$,
which expresses that $\pose$ is a single-valued function of $(s,t)$.
Approximating both derivatives by backward differences gives
\begin{equation}
    \frac{\vel_{j,k}-\vel_{j-1,k}}{\Delta s}
    = \frac{\strain_{j,k}-\strain_{j,k-1}}{\Delta t}
    - \curly{\strain_{j,k}}\vel_{j,k} ,
\label{eq:compat_bch}
\end{equation}
which we enforce through the residual
\begin{equation}
    \res^{\mathrm{c}}_{j,k}
    = \frac{\vel_{j,k}-\vel_{j-1,k}}{\Delta s}
    - \frac{\strain_{j,k}-\strain_{j,k-1}}{\Delta t}
    + \curly{\strain_{j,k}}\vel_{j,k} ,
    \label{eq:compat_factor}
\end{equation}
with associated cost $J^{\mathrm{c}}_{j,k}=\|\res^{\mathrm{c}}_{j,k}\|^{2}_{\bm{\Sigma}_{\mathrm{c}}^{-1}}$ on the interior nodes.

This is required because $\pose$, $\strain$, and $\vel$ are estimated separately: without it the optimizer may return a strain and velocity corresponding to no realizable rod motion.
Our factor is similar to \cite{teetaert2025stochastic} but carries more terms for accuracy.

\subsection{Momentum Balance Factor}
\label{sec:momentum_factor}

Because~\eqref{eq:wrench_sum} retains the algebraic form of the static balance, this factor is \emph{unchanged} from the quasi-static case \cite{ferguson2026continuum} apart from the wrench at each node now being the full sum $\wrenchtot$.
Given discrete wrench loads at the node locations, the residual is
\begin{equation}
\begin{aligned}
    \res^{\stress}_{j,k}
    &=
    \Ad^T\!\big(\pose_{j,k}^{-1}\pose_{j+1,k}\big)\,
    \stiff(\strain_{j,k} - \strain_0)
    \\
    &-\blockrot^T(\pose_{j+1,k})\,\wrenchtot_{j+1,k}
  -\stiff\strain_{j+1,k},
\label{eq:momentum_factor}
\end{aligned}
\end{equation}
with $\blockrot(\pose)=\mathrm{diag}(\rot,\rot)$ and
$J^{\stress}_{j,k}=\|\res^{\stress}_{j,k}\|^{2}_{\bm{\Sigma}_{\stress}^{-1}}$.
The three terms are a balance of wrenches at node $j+1$: the internal stress carried forward from node $j$, transported by $\Ad^{T}$ so that its moment is taken about the new node's origin; the applied load there, rotated into the body frame; and the internal stress the next node itself carries.
At equilibrium these cancel, and any imbalance is exactly what the node's own load must supply.
This residual is a wrench, so $\bm{\Sigma}_{\stress}$ accounts for any load not attributed to a term of $\wrenchtot$; it is the one place a load can enter anonymously, so we set $\bm{\Sigma}_{\stress}$ tight.

\subsection{Tendon Actuation}
\label{sec:actuation}

We follow prior work \cite{ferguson2026continuum} to map actuator variables $\brm{q}_k$ to the actuation wrenches $\wrenchact_{j,k}$ through an actuation model.
To account for actuator uncertainty, we model $\brm{q}_k$ as a random vector with Gaussian prior
\begin{equation}
\res^{q}_{k}=\brm{q}_{k}-\bar{\brm{q}}_{k},
\qquad
J^{q}_{k}=\|\res^{q}_{k}\|^{2}_{\bm{\Sigma}_{q}^{-1}},
\label{eq:actuation_prior}
\end{equation}
where $\bar{\brm{q}}_{k}$ denotes the measured actuator values at time $t_k$
and $\bm{\Sigma}_{q}$ the associated uncertainty.

Tendon routing discs are located at a subset
$\discset\subseteq\{1,\dots,N_s\}$ of arclength nodes. 
At non-disc nodes the actuation wrench vanishes, $\wrenchact_{j,k}=\brm{0}$, so we drop the variable in those locations. 
At a disc node $d\in\discset$, each tendon $i$ exerts wrenches $\wrench^{-}_{d,i,k}$ and
$\wrench^{+}_{d,i,k}$ from the two adjacent discs, taken as zero at the rod
ends. These follow from the tension $q_{i,k}$ and the relative routing hole
locations,
\begin{equation}
\wrench^{\pm}_{d,i,k}=\blockrot(\pose_{d,k})
  \begin{bmatrix}\hatop{\brm{h}_{d,i}}\\ \brm{I}_3\end{bmatrix}
  \tilde{\wrench}^{\pm}_{d,i,k},
\qquad
\tilde{\wrench}^{\pm}_{d,i,k}=q_{i,k}
  \frac{\brm{d}^{\pm}_{d,i,k}}{\|\brm{d}^{\pm}_{d,i,k}\|},
\label{eq:tendon_wrench}
\end{equation}
\begin{equation}
\brm{d}^{\pm}_{d,i,k}=\begin{bmatrix}\brm{I}_3 & 0\end{bmatrix}
  \pose_{d,k}^{-1}\pose_{d^{\pm},k}
  \begin{bmatrix}\brm{h}_{d^{\pm},i}\\ 1\end{bmatrix}-\brm{h}_{d,i},
\label{eq:hole_diff}
\end{equation}
where $\brm{h}_{d,i}$ is the body-frame hole location of tendon $i$ on disc
$d$, and $d^{\pm}$ index the adjacent discs. The normalized difference in hole
locations gives the direction of the tendon force, scaled by the tension.
Summing over tendons yields the actuation factor
\begin{equation}
\res^{\mathrm{a}}_{d,k}=\sum_{i=1}^{N}
  \big(\wrench^{-}_{d,i,k}+\wrench^{+}_{d,i,k}\big)-\wrenchact_{d,k},
\qquad
J^{\mathrm{a}}_{d,k}=\|\res^{\mathrm{a}}_{d,k}\|^{2}_{\bm{\Sigma}_{\mathrm{a}}^{-1}},
\label{eq:actuation_factor}
\end{equation}
with $\bm{\Sigma}_{\mathrm{a}}$ capturing small unmodeled effects such as
tendon--disc friction and routing imperfections. This links the tension
variables to the backbone wrenches at each time step.

\subsection{Inertial Wrench Factor}
\label{sec:inertial}

\begin{figure}[t]
    \centering
    \vspace{-20pt}
    \includegraphics[width=1.0\linewidth, clip, trim=600pt 600pt 100pt 0pt]{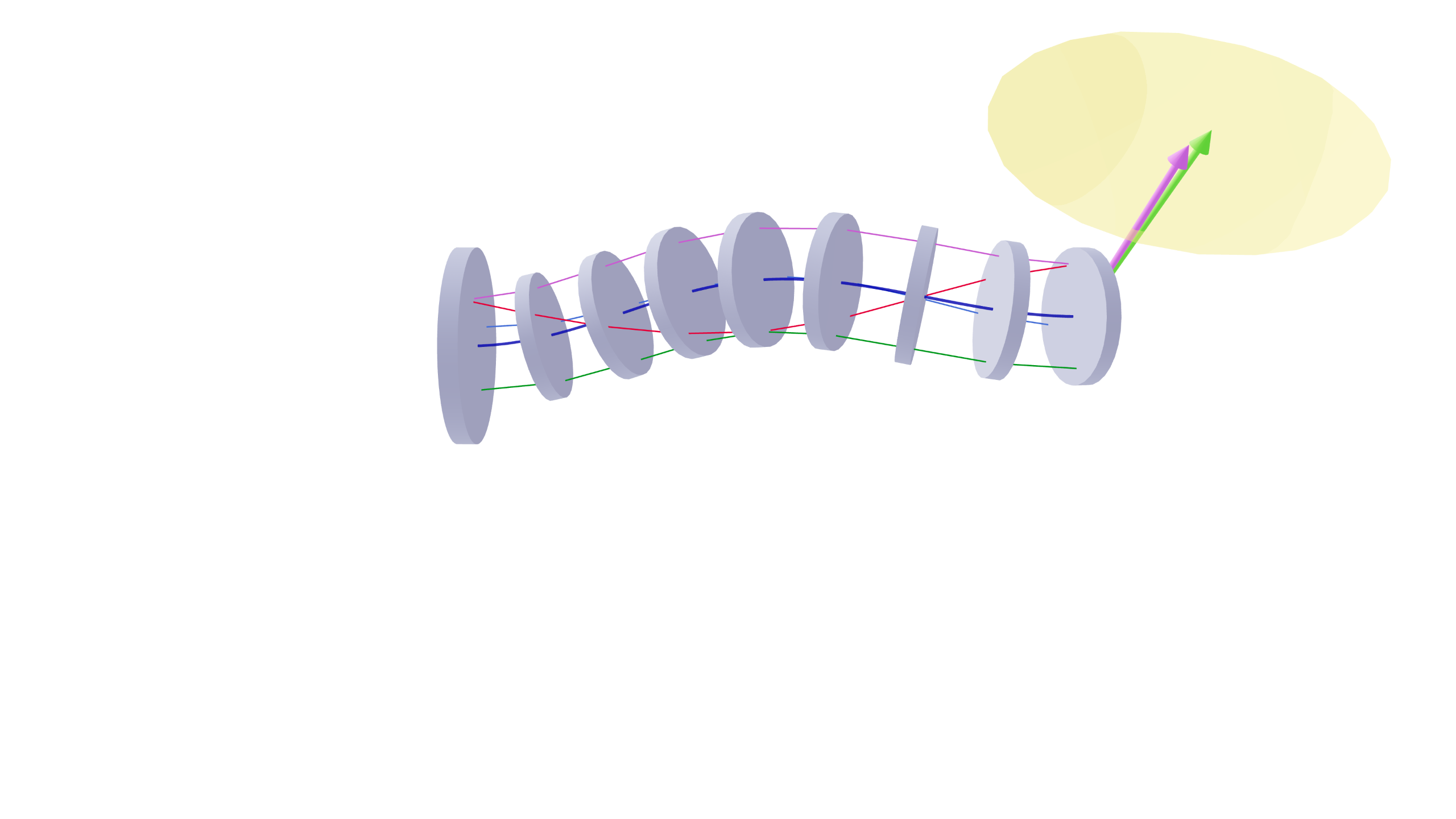}
    \includegraphics[width=0.9\linewidth, clip, trim=5pt 0pt 0pt 0pt]{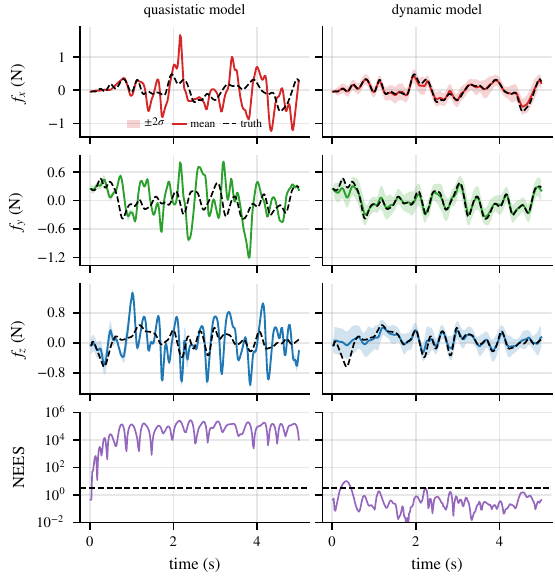}
    \vspace{-20pt}
    \caption{Tip force estimation in simulation, from noisy tension and tip position measurements. \textit{Top}: a snapshot, with estimated (magenta) and true (green) force and the $2\sigma$ uncertainty in gold. \textit{Bottom}: quasi-static \cite{ferguson2026continuum} (\textit{left}) against ours (\textit{right}), same trajectory and measurements. Rows give the force components---mean, $2\sigma$ band, truth (black, dashed)---then NEES against $\mathrm{E}[\chi^2_3]=3$ (dashed). The quasi-static NEES sits orders of magnitude high: wrong and confident at once. RMS relative to truth was \textbf{152 mN} for ours vs \textbf{733 mN} for the quasi-static solver. Solve times were \textbf{44.0} vs. \textbf{2.1} ms.}
    \label{fig:tendon_robot_sim}
    \vspace{-10pt}
\end{figure}

The inertial wrench $\wrenchin$ of~\eqref{eq:inertial_wrench} is the load a node exerts on the rod by virtue of its own acceleration, which this factor computes on the discrete grid.
The distributed mass of the segment around each node is lumped into the constant generalized inertia $\inertia$ of~\eqref{eq:dynamic_balance}; for the tendon robot the nodes coincide with the spacer discs, so $\inertia$ collects each disc, its markers, and the adjacent backbone into a single cylinder.
We express each node's momentum in world axes at its own step,
\begin{equation}
\momentum_{j,k} = \blockrot(\pose_{j,k})\,\inertia\,\vel_{j,k},
\label{eq:momentum}
\end{equation}
reusing $\blockrot$ from~\eqref{eq:momentum_factor}.
With this,
\begin{equation}
    \wrenchin_{j} = -\dot{\momentum}_{j} - \blockrot(\pose_{j})\,\damping\,\vel_{j} .
    \label{eq:inertial_wrench_defn}
\end{equation}
We discretize $\dot{\momentum}$ with the second-order backward differentiation formula (BDF2),
$\dot{\momentum}_{j,k} \approx (\tfrac{3}{2}\momentum_{j,k} - 2\momentum_{j,k-1} + \tfrac{1}{2}\momentum_{j,k-2})/\Delta t$.
BDF2 is L-stable and stiffly accurate, which matters because the shear and elongation modes of a stiff rod are far faster than the bending modes of interest \cite{till2019dynamics}.
Being a backward formula, it also requires time steps the moving window already holds.
Substituting~\eqref{eq:momentum} and evaluating the damping at $t_k$ gives
\begin{equation}
\begin{aligned}
    \res^{\mathrm{i}}_{j,k}
    &= \wrenchin_{j,k}
    + \blockrot(\pose_{j,k})\Big(\damping + \tfrac{3}{2\Delta t}\inertia\Big)\vel_{j,k}
    \\
    &- \tfrac{2}{\Delta t}\,\blockrot(\pose_{j,k-1})\,\inertia\,\vel_{j,k-1}
    + \tfrac{1}{2\Delta t}\,\blockrot(\pose_{j,k-2})\,\inertia\,\vel_{j,k-2},
\end{aligned}
    \label{eq:inertial_factor}
\end{equation}
with cost $J^{\mathrm{i}}_{j,k} = \|\res^{\mathrm{i}}_{j,k}\|^{2}_{\bm{\Sigma}_{\mathrm{i}}^{-1}}$.
Here $\bm{\Sigma}_{\mathrm{i}}$ accounts for error in the momentum balance at a node, chiefly an inexact lumped mass and damping that is not truly linear (e.g., static friction).
Being a two-step method, the factor is added from $k\ge2$; the two initial steps instead carry the velocity prior of~\eqref{eq:initial_velocity}.

Note that this factor, along with the temporal kinematics factor \eqref{eq:temporal_kin_factor}, overall links the individual spatial graphs (Fig.~\ref{fig:robot_graph}) across timesteps to form the full graph in Fig.~\ref{fig:intro_figure}.
As $\inertia$ and $\damping$ approach zero, \eqref{eq:inertial_factor} reduces to the quasi-static case \cite{ferguson2026continuum}; i.e., $\res^{\mathrm{i}}_{j,k} = \wrenchin_{j,k}$; essentially a tight zero prior on the inertial wrench.

\subsection{Boundary Conditions}
\label{sec:boundary}

The spatial chain at each time step is anchored at its two ends, by a prior on the base pose and one on the external load (Section~\ref{sec:wrench_prior}), as drawn in Fig.~\ref{fig:robot_graph}.
We anchor the base pose as in \cite{ferguson2026continuum},
\begin{equation}
\res^{\mathrm{p}}_{k}=\ln\big(\bar{\pose}^{-1}\pose_{1,k}\big)^{\vee},
\qquad
J^{\mathrm{p}}_{k}=\|\res^{\mathrm{p}}_{k}\|^{2}_{\bm{\Sigma}_{\mathrm{p}}^{-1}},
\label{eq:base_prior}
\end{equation}
with $\bm{\Sigma}_{\mathrm{p}}$ accounting for mounting and registration error, applied identically at every time step.

The temporal kinematics factor~\eqref{eq:temporal_kin_factor} couples two consecutive
time steps and the inertial wrench factor~\eqref{eq:inertial_factor} three, so the first two steps require separate treatment. 
For $k\in\{1,2\}$ no inertial wrench variable is introduced and the wrench sum reduces to $\wrenchtot=\wrenchext+\wrenchact+\wrenchgrav$, so the graph at those steps is exactly the quasi-static model of prior work\cite{ferguson2026continuum}, initializing the rod at static equilibrium under gravity and actuation. 
The velocities are constrained by a zero-mean prior
\begin{equation}
\res^{v_0}_{j,k}=\vel_{j,k},
\qquad
J^{v_0}_{j,k}=\|\res^{v_0}_{j,k}\|^{2}_{\bm{\Sigma}_{v_0}^{-1}},
\label{eq:initial_velocity}
\end{equation}
appropriate for a robot starting from rest, and giving BDF2 the two consistent static states it needs before dynamics can start.

Because each time step adds a fixed number of variables and factors, the graph would grow without bound along a trajectory.
We therefore solve over a sliding window~\cite{teetaert2026sliding} using a fixed-lag smoother~\cite{dellaert2012factor}, retaining the most recent $L$ time steps and marginalizing the rest into a \emph{state history prior} over the oldest retained step (Section~\ref{sec:windowed}).
That step is therefore constrained by this marginal rather than by~\eqref{eq:initial_velocity}; the initial-condition factors apply only at the true start of the trajectory.

\subsection{External Wrench Prior}
\label{sec:wrench_prior}

Where no external load acts, no variable is introduced and the term is simply absent from $\wrenchtot$.
Where one does, it enters as a variable with a Gaussian prior
\begin{equation}
\res^{\mathrm{e}}_{k}=\wrenchext_{k}-\bar{\wrench}^{\mathrm{e}}_{k},
\qquad
J^{\mathrm{e}}_{k}=\|\res^{\mathrm{e}}_{k}\|^{2}_{\bm{\Sigma}_{\mathrm{e}}^{-1}},
\label{eq:wrench_prior}
\end{equation}
whose role depends on $\bm{\Sigma}_{\mathrm{e}}$: a tight prior states a known load, as when a wrench is commanded in simulation. A loose prior leaves the load to be inferred from measurements.
In both cases we model contact as a point force and keep the moment block of $\bm{\Sigma}_{\mathrm{e}}$ small.

\subsection{Measurement Factors}
\label{sec:measurements}

We use two kinds of observations in our experiments. 
The first is a position measurement, as in prior work~\cite{ferguson2026continuum}: if $\brm{z}_{j,k}$ is an observed position of node $j$ at $t_k$, the Gaussian residual is
\begin{equation}
    \res^{\mathrm{z}}_{j,k} =
    \brm{z}_{j,k} - \mathrm{Pos} (\pose_{j,k}),
\label{eq:meas_factor}
\end{equation}
contributing $J^{\mathrm{z}}_{j,k} = \|\res^{\mathrm{z}}_{j,k}\|^{2}_{\bm{\Sigma}_{\mathrm{z}}^{-1}}$ with $\bm{\Sigma}_{\mathrm{z}}$ the sensor noise.
Such measurements are what make load estimation well posed \cite{aloi2022estimating, rucker2011deflection, ferguson2024unified, ferguson2026continuum}.

The second type of measurement is a fiber Bragg grating (FBG) shape sensor \cite{shi2016shape, xu2016curvature, modes2020shape}, whose readings along the fiber give the two bending components of the strain at a node.
With $\brm{z}^{\strain}_{j,k}$ the measurement, the residual is
\begin{equation}
\res^{\kappa}_{j,k}=\brm{z}^{\strain}_{j,k}-\begin{bmatrix}\brm{I}_2 & \brm{0}\end{bmatrix}\strain_{j,k},
\qquad
J^{\kappa}_{j,k}=\|\res^{\kappa}_{j,k}\|^{2}_{\bm{\Sigma}_{\kappa}^{-1}},
\label{eq:curvature_factor}
\end{equation}
with $\bm{\Sigma}_{\kappa}$ the fiber's stated noise.
Note that \eqref{eq:curvature_factor} constrains the backbone's \emph{shape} without observing its position in space, which makes it complementary to~\eqref{eq:meas_factor}.

\subsection{Windowed Optimization}
\label{sec:windowed}

Under the Gaussian noise assumptions above, the negative log posterior is proportional to a sum of many weighted squared residuals of the form
\begin{equation}
    J_{\mathrm{tot}} = \sum_i \| \res_i \|^2_{\Sigma_i^{-1}},
\label{eq:map_objective}
\end{equation}
which we do not write out explicitly, favoring instead the equivalent factor graph representation of Figs.~\ref{fig:intro_figure} and \ref{fig:robot_graph} \cite{dellaert2012factor, ferguson2026continuum}.
The sum is overall combines all factors into a single cost to minimize for the most likely trajectory over the window.

The cost of minimizing~\eqref{eq:map_objective} over the full trajectory grows
without bound, so we instead solve it over a sliding window~\cite{teetaert2026sliding}
using a batch fixed-lag smoother built on GTSAM~\cite{dellaert2012factor}.
At each step, the new variables and factors are inserted, the window is optimized
with Powell's dogleg method, warm started from the previous estimate, and variables
older than a lag $\lag$ behind the current time are marginalized out.
Before any variables have been marginalized, a prior on the initial state anchors
the window.

Marginalization summarizes the information carried by the aged-out variables
$\mathcal{X}_{\mathrm{old}}$ as a Gaussian prior on $\mathcal{S}$, the remaining
variables that share a factor with them.
The factors touching $\mathcal{X}_{\mathrm{old}}$ are linearized, and
$\mathcal{X}_{\mathrm{old}}$ is eliminated from this subgraph, giving the negative log-likelihood
\begin{equation}
    -\tfrac{1}{2}
  \big\|\brm{A}_{\mathcal{S}}\,\delta\mathcal{S}-\brm{b}_{\mathcal{S}}\big\|^{2}
\label{eq:marginal_factor}
\end{equation}
where $\delta\mathcal{S}$ is a perturbation about the linearization point.
This prior replaces the eliminated factors. The trailing edge of the window
therefore retains the information from all past measurements, exactly up to
linearization, rather than starting uninformed.

Importantly, the prior does not densely couple the window.
Eliminating a variable creates fill-in only among its neighbors, so
\eqref{eq:marginal_factor} involves only $\mathcal{S}$. Within $\mathcal{S}$, it
couples only variables that are connected through $\mathcal{X}_{\mathrm{old}}$.
GTSAM performs this elimination automatically and keeps the result in factored
form: each connected group of marginalized variables yields its own factor, which
is dense only over that group's separator.
Because aged-out states connect to the window through only a few neighboring
variables, the prior stays small and the system remains sparse.

We take the Laplace approximation at the MAP solution as the posterior, so the covariance is the inverse Gauss--Newton Hessian at the optimum; forming it densely would be prohibitive, but marginals over any subset of variables follow from the same sparse elimination used for the solve.
For vector-valued states the marginal is an ordinary Gaussian; for poses it is
expressed in the tangent space at the estimate, as a mean $\bar{\pose}_{j,k}\in\SE$ with covariance $\bm{\Sigma}_{\pose_{j,k}}\in\mathbb{R}^{6\times6}$ over the local perturbation
\begin{equation}
\pose_{j,k}=\bar{\pose}_{j,k}\exp\!\big(\hatop{\bm{\xi}_{j,k}}\big),
\qquad
\bm{\xi}_{j,k}\sim\mathcal{N}\big(0,\bm{\Sigma}_{\pose_{j,k}}\big).
\label{eq:pose_marginal}
\end{equation}
The Gaussian lives in $\se$ rather than on the group.

\section{Simulations}
\label{sec:sim}

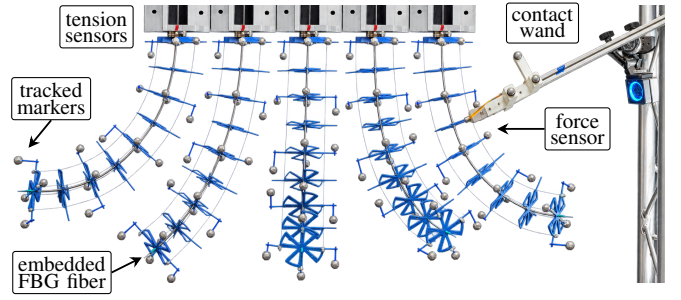
\begin{figure}
    \centering
    \input{figures/experiment_setup.tikz}
    \caption{Experimental setup from the open dataset, adapted from Gotelli et al.~\cite{gotelli2026multi}. A tendon-driven continuum robot is driven through dynamic, oscillatory trajectories while recording tendon tensions, motion capture of the discs, and an FBG shape-sensing fiber. All preprocessing was done by the authors of~\cite{gotelli2026multi}.}
    \label{fig:experiments}
\end{figure}

\subsection{Simple Cosserat Rod Dynamics}
\label{sec:sim_rod}

We first check that the model reproduces Cosserat rod dynamics with no measurements at all (i.e. stochastic forward simulation).
All actuation variables and factors are removed, gravity and damping are set to zero, and the only input per step is the tip wrench prior of Section~\ref{sec:wrench_prior}.
A moment and force are held at the tip until $0.1$~s and then released, after which the rod rings freely.
We compare against Till et al.~\cite{till2019dynamics}, whose scheme discretizes in time and solves the resulting arclength boundary value problem by RK4 and single shooting.
Results are shown in Fig.~\ref{fig:cosserat_rod_example}.

\subsection{Helically Routed Tendon Robot}
\label{sec:sim_tendon}

To evaluate tip force estimation, we simulate the robot of Section~\ref{sec:real} with one tendon rerouted helically, so that actuation induces torsion and out-of-plane bending.
Ground truth is generated with the same model, driven by prescribed tendon
tensions and a time-varying tip force.
Both the true force and the measurement noise are drawn from smooth Gaussian
processes, since real sensor errors are correlated in time rather than white.
The tension amplitude is chosen so that inertial forces are comparable in
magnitude to the tip force.
We run $250$ steps at $50$~Hz with $L=5$.

The estimator receives noisy tendon tension and tip position measurements and a
weak prior on the tip force. It must infer the force from the discrepancy between
the measured tip position and the position predicted from the tensions alone.
We compare against the quasi-static model of~\cite{ferguson2026continuum}, which
receives the same measurements but solves each frame independently.
Results are shown in Fig.~\ref{fig:tendon_robot_sim}.

\section{Real-World Evaluation}
\label{sec:real}

\begin{figure*}
    \centering
    \includegraphics[width=0.8\linewidth]{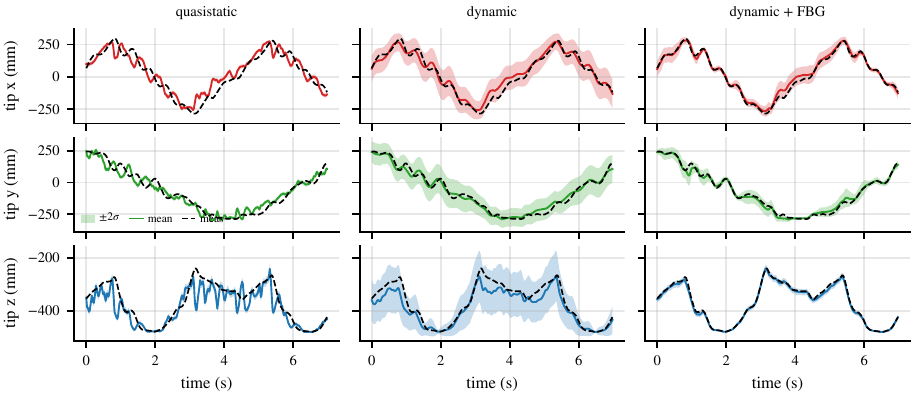}
    \hfill
    \input{figures/tip_pred_comparison.tikz}
    \vspace{-10pt}
    \caption{Shape estimation on the dataset of \cite{gotelli2026multi}. 
    \textit{Left}: Predicted tip position against motion capture (black, dashed) with the $2\sigma$ band, for the quasi-static baseline \cite{ferguson2026continuum}, our dynamic model with tensions only, and our dynamic model with the FBG data~\eqref{eq:curvature_factor}. \textit{Right}: the corresponding backbone estimates. RMS relative to measured position was \textbf{83.7 mm} for quasistatic vs. \textbf{48.4 mm} for the dynamic model vs. \textbf{25.6 mm} with FBG measurements. Solve times were \textbf{7.5} vs. \textbf{40.2} vs. \textbf{55.0} ms respectively.}
    \label{fig:results}
\end{figure*}

We apply our method directly to the open dataset of Gotelli et al. \cite{gotelli2026multi}, whose setup is shown in Fig.~\ref{fig:experiments}.
Their robot is a $0.48$~m backbone with nine discs, hanging tip-downward along gravity and driven by four tendons.
We use three of their time-aligned streams: tendon tensions; motion capture of five discs, which provides the tip position we score against; and an FBG fiber carrying $26$ gratings at $20$~mm spacing.
Disc spacing is $60$~mm, so every third grating coincides with a disc and~\eqref{eq:curvature_factor} applies at the seven interior nodes.
We run the smoother at $50$~Hz with $L = 5$.

\subsection{Calibration}
\label{sec:calib}

We fit the bending stiffness in $\stiff$ and the kinematics covariance $\bm{\Sigma}_{\strain}$ by maximum likelihood on the tip position over the quasi-static bending sweeps, and hand tune the damping $\damping$ and the dynamics covariance $\bm{\Sigma}_{\mathrm{i}}$ against an open-loop run on a dynamic recording.
We reuse $\bm{\Sigma}_{\strain}$ for $\bm{\Sigma}_{\vel}$, and take $\bm{\Sigma}_{\mathrm{z}}$, $\bm{\Sigma}_{\kappa}$ and the tension prior from the noise tables of \cite{gotelli2026multi}.
Parameter values and the full procedure are in the accompanying code.

\subsection{Shape Estimation}
\label{sec:real_shape}

Three estimators are run over an $8$~s window of the fast Lissajous trajectory of \cite{gotelli2026multi}: the quasi-static baseline, our dynamic model, and our dynamic model with the FBG factor~\eqref{eq:curvature_factor} at the interior nodes.
\emph{None of the three is given the tip position for estimation}, only the measured tendon tensions, plus the FBG measurements in the third case\new{.}
All runs are scored against motion capture at the tip in Fig.~\ref{fig:results}.

\section{Discussion}

With no measurements, our method acts as a pure forward simulation. It reproduces
the results of Till et al.~\cite{till2019dynamics} to within 3~mm RMS while
using fifteen times fewer spatial nodes (Fig.~\ref{fig:cosserat_rod_example}).
It also returns a covariance over the full rod state, which captures how
uncertainty in the inputs and model propagates through the dynamics.
We attribute the accuracy at coarse resolution largely to the Magnus expansion in
our kinematics factors~\eqref{eq:spatial_kin_factor} and~\eqref{eq:temporal_kin_factor},
which integrates the rod's pose between nodes to higher order.
Some phase drift accumulates over the trajectory, as expected given the different
discretizations; a fuller comparison with~\cite{till2019dynamics} is left to
future work.

The quasi-static estimator fails in both tendon robot experiments, and for the
same underlying reason: it has no model of inertia, so it must explain inertial
effects with the quantities it does model.
When given only tendon tension measurements (Fig.~\ref{fig:results}, left), the
robot's oscillations raise the measured tensions. The quasi-static model can
balance these tensions only with elastic deformation, so its estimated tip
position diverges from the ground truth (Fig.~\ref{fig:results}, middle), with an
RMS error of roughly double our method, which correctly attributes the rise in tension to the robot's
acceleration.
In the force estimation simulation (Fig.~\ref{fig:tendon_robot_sim}), the same
missing term corrupts the force estimate: oscillations in the measured tip
position that are caused by inertia are interpreted as external loading.
The resulting RMS force error is roughly fivefold ours. Because our model predicts the inertial wrench, it can separate
it from the external wrench applied at the tip.

In both experiments, the quasi-static model also fails \emph{confidently}.
Its $2\sigma$ bands remain tight around estimates that are far from the truth, and
in the force estimation simulation its NEES is orders of magnitude above the
expected value, whereas our estimator is consistent
(Fig.~\ref{fig:tendon_robot_sim}).
Overconfidence is arguably worse than inaccuracy alone, since downstream consumers
such as controllers or safety monitors receive no signal that the estimate should
be distrusted.

The shape estimation experiment shows that our method can estimate the rod's
shape both with and without FBG strain measurements.
With FBG data, the tip position RMS error is reduced by roughly a factor of two.
In both cases, the true tip position lies visually within the estimated $2\sigma$
region, so the reported uncertainty is reasonable as well as the estimate.

Our estimator took roughly $50$~ms per step, compared with roughly $5$~ms for the quasi-static solver.
Some of this gap can likely be closed through implementation improvements, but
much of it reflects a modeling trade-off.
When inertial effects are small relative to the applied loads, as in slow,
lightweight instruments such as many surgical robots, a quasi-static model may
suffice, and its speed is an advantage.
For larger or faster robots, where inertial forces are comparable to external
loads, modeling the dynamics is necessary for accurate estimation.
Note that more thorough analysis of the speed and complexity of this approach, especially with respect to lag window, will be the topic of future work.

Optimization was stable on all trajectories reported here, but not on every
trajectory we tested.
On a few aggressive trajectories the solver failed to converge, which we attribute
to poor conditioning.
This motivated both the stress parameterization of Section~\ref{sec:spatial_kin}
and the L-stable discretization of Section~\ref{sec:inertial}.
Determining the stability regime and comparing with quasi-static estimators will be the topic of future work. 

\section{Conclusion}

We have presented a factor graph formulation of continuum robot dynamics.
Solved over a sliding window, the same graph is a stochastic forward simulation when no measurements are available.
With measurements, the approach acts as a state estimator, reporting uncertainty that stays calibrated during motion.
We compared against a Cosserat dynamics benchmark \cite{till2019dynamics} for the forward simulation, and against a quasi-static factor graph estimator \cite{ferguson2026continuum} on both tendon robot experiments.
Because the solver operates on a factor graph, heterogeneous sensing enters as additional factors, which we demonstrated by fusing tip position measurements and an FBG shape sensor.
Loads of unknown or distributed location, and model-predictive control against a resulting posterior horizon, are natural next steps.

\section*{ACKNOWLEDGMENT}

We thank Gotelli et al.~\cite{gotelli2026multi} for making their comprehensive experimental dataset publicly available, which enabled the experiments presented in this work.

\bibliographystyle{IEEEtran}
\bibliography{library}

\end{document}

%% file: preamble.tex
\usepackage{graphicx}
\usepackage{cite}
\usepackage[dvipsnames]{xcolor}
\usepackage{amsmath,amssymb,bm,mathtools}
\usepackage{tikz}
\usetikzlibrary{arrows.meta}
\usetikzlibrary{fit}
\usetikzlibrary{calc}
\usetikzlibrary{fadings}
\usetikzlibrary{decorations.pathreplacing}
\usetikzlibrary{positioning}
\usepackage[urlcolor=blue]{hyperref}
\usepackage{multirow}
\usepackage{booktabs}

\definecolor{myblue}{RGB}{0, 71, 255}
\definecolor{myred}{RGB}{255, 43, 43}
\definecolor{myyellow}{RGB}{255, 255, 100} 
\definecolor{mygreen}{RGB}{115,208,115} 
\definecolor{mygray}{RGB}{200, 200, 200}
\definecolor{myorange}{RGB}{234,134,66}
\definecolor{mypurple}{RGB}{155,0,255}

\newcommand{\new}[1]{\textcolor{blue}{#1}}

\newcommand{\brm}[1]{\bm{\mathrm{#1}}}
\newcommand{\pose}{\brm{T}}
\newcommand{\vel}{\brm{v}}
\newcommand{\rot}{\brm{R}}
\newcommand{\pos}{\brm{p}}
\newcommand{\curly}[1]{{#1}^{\curlywedge}}
\newcommand{\curlyT}[1]{{#1}^{{\curlywedge}^T}}
\newcommand{\hatop}[1]{{#1}^{\wedge}}
\newcommand{\magnus}{\bm{\phi}}
\newcommand{\momentum}{\bm{\mu}}
\newcommand{\blockrot}{\bm{\mathcal{R}}}
\newcommand{\wrenchext}{\wrench^{\mathrm{e}}}
\newcommand{\wrenchact}{\wrench^{\mathrm{a}}}
\newcommand{\wrenchgrav}{\wrench^{\mathrm{g}}}
\newcommand{\wrenchin}{\wrench^{\mathrm{i}}}
\newcommand{\wrenchtot}{\wrench^{\mathrm{tot}}}
\newcommand{\res}{\brm{e}}
\newcommand{\discset}{\mathcal{P}}
\newcommand{\stress}{\brm{\sigma}}
\newcommand{\strain}{\brm{\epsilon}}
\newcommand{\wrench}{\brm{f}}

\newcommand{\Ad}{\operatorname{Ad}}

\newcommand{\se}{\mathfrak{se}(3)}
\newcommand{\SE}{\mathrm{SE}(3)}
\newcommand{\inertia}{\brm{M}}
\newcommand{\damping}{\brm{D}}
\newcommand{\stiff}{\brm{K}}

\newcommand{\lag}{L}

\tikzset{
    state/.style={
        draw,
        circle,
        fill=myblue,
        minimum size=0.16cm,
        inner sep=0
    },
    cosserat_factor/.style={
        draw,
        fill=myred,
        minimum size=0.13cm,
        inner sep=0,
        rectangle
    },
    yellow_factor/.style={
        draw,
        fill=myyellow,
        minimum size=0.13cm,
        inner sep=0,
        rectangle
    },
    purple_factor/.style={
        draw,
        fill=mypurple,
        minimum size=0.13cm,
        inner sep=0,
        rectangle
    },
    prior_factor/.style={
        draw,
        fill=mygreen,
        minimum size=0.13cm,
        inner sep=0,
        rectangle
    },
    callout/.style={
        align=center,
        draw,
        rounded corners=1pt,
        inner sep=2pt,
        font=\footnotesize\linespread{0.7}\selectfont
    },
    callout arrow/.style={
        -{Stealth[length=1.5mm,width=1.5mm]},
        line width=0.7pt
    },
    disc/.style={draw=none, fill=black!10, minimum width=7.0mm, minimum height=13mm, inner sep=0pt},
}

%% file: figures/front_page.tikz
\begin{tikzpicture}[scale=1.0, every node/.style={font=\footnotesize}]
    \node[] at (2.8, 3.0) {
        \includegraphics[width=0.9\linewidth, height=0.47\linewidth]{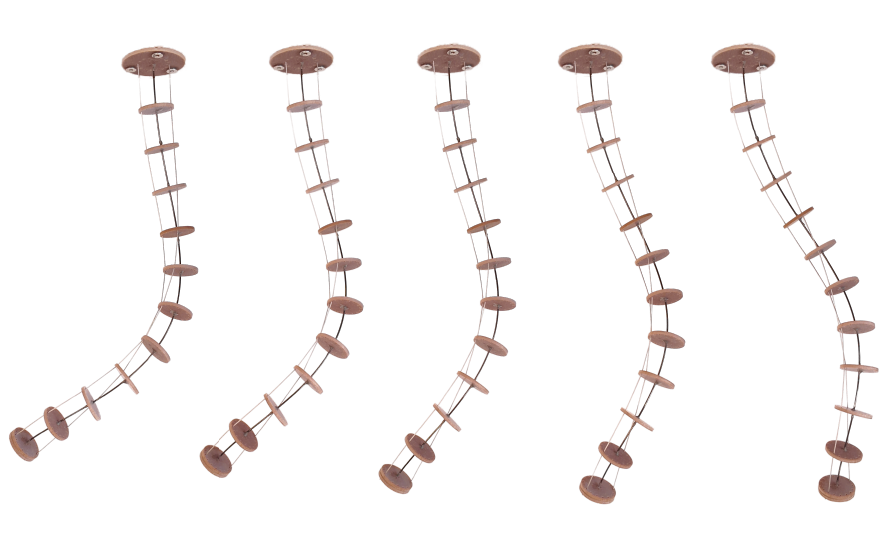}
    };

    \def\ds{1.1}   
    \def\dt{1.5}   

    \foreach \j in {0,...,4}{
        \node[draw=none, fill=black!10, rectangle, inner xsep=0.4cm, inner ysep=1.85cm]
            (B\j) at (\j*\dt, -1.5*\ds) {};

        \node[draw=none, fill=black!40, rectangle, inner xsep=0.3cm, inner ysep=0.05cm]
            at (\j*\dt, 0.025) {};

        \draw[line width=3, color=black!40]
            plot[domain=0:3, samples=30, smooth]
            ({\j*\dt - 0.04*\x*\x + 0.15*(1-cos(\x/3*230))}, {-\x*\ds});
        
        \draw[]
            plot[domain=0:3, samples=30, smooth]
            ({\j*\dt - 0.04*\x*\x + 0.15*(1-cos(\x/3*230))}, {-\x*\ds});
        
        \foreach \i in {0,...,3}{
            \node[state, fill=myblue] (S\i\j) at ({\j*\dt - 0.04*\i*\i + 0.15*(1-cos(\i/3*230))}, -\i*\ds) {};
        }

        \foreach \a in {0, 1, 2}{
            \pgfmathsetmacro{\i}{\a+0.5}
            \node[cosserat_factor] (F\a\j) at ({\j*\dt - 0.04*\i*\i + 0.15*(1-cos(\i/3*230))}, -\i*\ds) {};
        }
    }
    
    \foreach \i in {1,...,3}{
    \foreach \j in {0,...,3}{
        \pgfmathtruncatemacro{\jp}{\j+1}
        \pgfmathtruncatemacro{\jm}{\j-1}

        \pgfmathsetmacro{\ypos}{-\ds*\i + 0.0)}

        \node[yellow_factor] (D\i\j)
            at (\dt*\j + 0.5*\dt, \ypos) {};

        \draw (S\i\j) -- (D\i\j) -- (S\i\jp);

        \def\vp{0.4}
        \ifnum\j>0
        \ifodd\j
            \draw (S\i\jm)
                .. controls +(\vp,\vp) and +(-\vp,\vp)
                .. (D\i\j);
        \else
        \draw (S\i\jm)
            .. controls +(\vp,\vp) and +(-\vp,\vp)
            .. (D\i\j);
        \fi
        \fi
    }
}

    \node[anchor=north, align=center] at ($(0, -3*\ds - 0.35)$) {$t_{K-L}$};
    \node[anchor=north, align=center] at ($(4*\dt, -3*\ds - 0.35)$) {$t_K$};
    \node[anchor=north, align=center] at ($(3*\dt, -3*\ds - 0.35)$) {$t_{K-1}$};
    \node[anchor=north, align=center] at ($(2*\dt, -3*\ds - 0.35)$) {$t_{K-2}$};
    \node[anchor=north, align=center, scale=1.5] at ($(1*\dt, -3*\ds - 0.35)$) {$\hdots$};

    \def\yc{0.7}
    
    \node[align=center] (A) at ($(3.5*\dt, \yc)$) {temporal and\\dynamics factors};
    \coordinate (C) at (3.5*\dt, 0.2);
    \draw[callout arrow] (C) -- ($(D13)!0.25!(C)$);

    \node[align=center] (A) at ($(2*\dt, \yc)$) {spatial\\factors};
    \draw[callout arrow] (A.south east) -- ($(F03)!0.1!(A)$);
    \draw[callout arrow] (A.south west) -- ($(F01)!0.1!(A)$);
    
    \node[align=center] (A) at ($(-0.15*\dt, \yc)$) {spatial, actuation, and\\meas. factors at $t_k$};

    \def\bw{0.4}
    \def\by{0.25}
    \draw[decorate, decoration={brace, amplitude=4pt}, line width=0.8pt] ($(1*\dt - \bw, \by)$) -- ($(1*\dt + \bw, \by)$);
    \coordinate (C) at ($(1*\dt, \yc)$);
    \draw[line width=0.7pt] (A.east) -- (C);
    \draw[line width=0.7pt] (C) -- ($(1*\dt, \yc - 0.15)$);

    \node[anchor=west] at ($(4*\dt, 0) + (0.5, 0)$) {$s_0$};
    \node[anchor=west] at ($(4*\dt, -\ds) + (0.5, 0)$) {$s_1$};
    \node[anchor=west, scale=1.5] at ($(4*\dt, -2*\ds) + (0.5, 0.2)$) {$\vdots$};
    \node[anchor=west] at ($(4*\dt, -3*\ds) + (0.5, 0)$) {$s_N$};

    \node[prior_factor] (M) at ($(-0.95,-1.25*\ds)$) {};
    \node[anchor=south, align=center] at (M.north) {state\\history\\prior};
    \foreach \i in {1,...,3}{
        \draw (M) -- (S\i0);
    }

    \coordinate (T0) at (-0.5, 5.1);
    \def\Tlen{7.0cm}

    \draw[-{Stealth[length=2mm,width=3mm]}, line width=0pt, color=BrickRed]
    (T0) -- ($(T0)+(\Tlen,0)$);
    \shade[left color=BrickRed!0, right color=BrickRed!100]
    ($(T0)+(0,-1pt)$) rectangle ($(T0)+(\Tlen-1mm,1pt)$);

    \node[callout, fill=white] at ($(T0)+(0.78*\Tlen,0)$) {time};

    \coordinate (S0) at (-1.1, 4.7);
    \def\Slen{2.1cm}
    
    \draw[-{Stealth[length=2mm,width=3mm]}, line width=0pt, color=BrickRed]
    (S0) -- ($(S0)+(0,-\Slen)$);
    \shade[top color=BrickRed!0, bottom color=BrickRed!100]
    ($(S0)+(-1pt,0)$) rectangle ($(S0)+(1pt,-\Slen + 1mm)$);
    \node[callout, fill=white] at ($(S0)+(0,-0.6*\Slen)$) {space};

    \def\y{2.8}
    \def\len{0.4cm}
    
    \coordinate (A) at (0.8,\y);
    \draw[-{Stealth[length=1.8mm,width=2.5mm]}, line width=1.5pt, color=black]
    (A) -- ($(A)+(\len, 0)$);

    \coordinate (A) at (2.4,\y);
    \draw[-{Stealth[length=1.8mm,width=2.5mm]}, line width=1.5pt, color=black]
    (A) -- ($(A)+(\len, 0)$);

    \coordinate (A) at (3.9,\y);
    \draw[-{Stealth[length=1.8mm,width=2.5mm]}, line width=1.5pt, color=black]
    (A) -- ($(A)+(\len, 0)$);
    
    \coordinate (A) at (5.5,\y);
    \draw[-{Stealth[length=1.8mm,width=2.5mm]}, line width=1.5pt, color=black]
    (A) -- ($(A)+(\len, 0)$);
\end{tikzpicture}

%% file: figures/robot_graph.tikz
\begin{tikzpicture}[scale=1.0, every node/.style={font=\footnotesize}]
    \def\dy{0.7}
    \def\dx{0.95}
    \def\amp{0.4}          
    \pgfmathsetmacro{\Xmax}{2*8*\dx}
    \pgfmathsetmacro{\PI}{3.14159265}

    \tikzset{
        S_label/.style={below, yshift=-2pt},
        T_label/.style={below, yshift=-2pt},
        F_label/.style={left, xshift=-2pt}
    }

    \draw[black!30, line width=4pt]
        plot[domain=0:\Xmax, samples=100, smooth]
        ({\x}, {\amp*(1-cos(\x/\Xmax*180))});

    \foreach \i in {3,5,7,9}{
        \pgfmathsetmacro{\th}{(\i-1)/8*180}
        \pgfmathsetmacro{\xx}{(\i-1)*2*\dx}
        \pgfmathsetmacro{\yy}{\amp*(1-cos(\th))}
        \node[disc] at (\xx,\yy) {};
    }
    
    \draw plot[domain=0:\Xmax, samples=100, smooth]
        ({\x}, {\amp*(1-cos(\x/\Xmax*180))});

    \node[draw=none, fill=black!10, rectangle, inner xsep=0.2cm, inner ysep=1.0cm] at (-0.2cm,0) {};

    \node[state] (T1) at (0,0) {};
    \node[below, yshift=22pt, xshift=12pt] at (T1) {$\brm{T}_{1,k}$};
    \node[state] (S1) at (0.0, -2.0*\dy) {};
    \node[below, yshift=-2pt] at (S1) {$\strain_{1,k}$};
    \foreach \i in {2,...,9}{
        \pgfmathsetmacro{\th}{(\i-1)/8*180}
        \pgfmathsetmacro{\yy}{\amp*(1-cos(\th))}
        \node[state] (T\i) at ({(\i-1)*2*\dx}, \yy) {};
        \node[T_label] at (T\i) {$\brm{T}_{\i, k}$};
        \node[state] (S\i) at ($(T\i)+(0,-2.0*\dy)$) {};

        \ifnum\i=9
            \node[S_label] at (S\i) {$\wrench^\mathrm{e}_k$};
        \else
            \node[S_label] at (S\i) {$\strain_{\i, k}$};
        \fi
    }

    \foreach \j [evaluate=\j as \k using int(\j+1)] in {1,...,8}{
        \node[cosserat_factor] (E\j) at ($(T\j)!0.5!(T\k)$) {};
        \draw (S\j) -- (E\j);
        \draw (S\k) -- (E\j);

        \node[state] (A) at ($(E\j) + (0,-1.1*\dy)$) {};
        \draw (A) -- (E\j);
        \node[anchor=north] at (A.south) {$\wrench^\mathrm{tot}_{j, k}$};
    }

    \foreach \dn/\cn in {1/3, 2/5, 3/7, 4/9}{
        \node[yellow_factor] (D\dn) at ($(T\cn)+(0,1.5)$) {};
    }
    \draw (T1) -- (D1); \draw (T3) -- (D1); \draw (T5) -- (D1);
    \draw (T3) -- (D2); \draw (T5) -- (D2); \draw (T7) -- (D2);
    \draw (T5) -- (D3); \draw (T7) -- (D3); \draw (T9) -- (D3);
    \draw (T7) -- (D4); \draw (T9) -- (D4);
  
    \def\fx{-1.5*\dx}
    \node[state] (F1) at ($(D1) + (\fx,0)$) {};
    \node[F_label, anchor=east] at (F1.west) {$\brm{f}^\text{a}_{3,k}$};
    \draw (D1) -- (F1);
    
    \node[state] (F2) at ($(D2) + (\fx,0)$) {};
    \node[F_label, anchor=east] at (F2.west) {$\brm{f}^\text{a}_{5,k}$};
    \draw (D2) -- (F2);
    
    \node[state] (F3) at ($(D3) + (\fx,0)$) {};
    \node[F_label, anchor=east] at (F3.west) {$\brm{f}^\text{a}_{7,k}$};
    \draw (D3) -- (F3);
    
    \node[state] (F4) at ($(D4) + (\fx,0)$) {};
    \node[F_label, anchor=east] at (F4.west) {$\brm{f}^\text{a}_{9,k}$};
    \draw (D4) -- (F4);

    \def\yrail{3.9*\dy} 
    \coordinate (railref) at (0,\yrail);
    \foreach \dn/\cn in {1/3, 2/5, 3/7, 4/9}{
        \coordinate (Q\dn) at (D\dn |- railref);
        \draw (D\dn) -- (Q\dn);
    }
    \draw (Q1) -- (Q2) -- (Q3) -- (Q4);

    \coordinate (qpos) at (1,\yrail);
    \node[state] (q) at (qpos) {};
    \draw (q) -- (Q1);
    \node[yshift=-6pt] at (q.south) {$\brm{q}_k$};

    \def\px{1.0*\dx}

    \node[prior_factor] (BP) at (-\px,0) {};
    \node[anchor=north, align=center] at (BP.south) {base\\pose\\prior};
    \draw (T1) -- (BP);
    
    \node[prior_factor] (TM) at ($(q)+(-\px,0)$) {};
    \node[anchor=north, align=center] at (TM.south) {tension\\meas.};
    \draw (q) -- (TM);

    \node[prior_factor] (TR) at ($(T9)+(\px,0)$) {};
    \node[anchor=south, align=center] at (TR.north) {tracker\\meas.};
    \draw (T9) -- (TR);

    \node[prior_factor] (TL) at ($(S9)+(\px,0)$) {};
    \node[anchor=north, align=center] at (TL.south) {tip load\\prior};
    \draw (S9) -- (TL);
\end{tikzpicture}

%% file: figures/experiment_setup.tikz
\begin{tikzpicture}[scale=1.0, every node/.style={font=\footnotesize}]
    \node[inner sep=0pt, anchor=center] (img) {
        \includegraphics[width=1.0\linewidth, clip, trim=1cm 1cm 2cm 0.5cm]{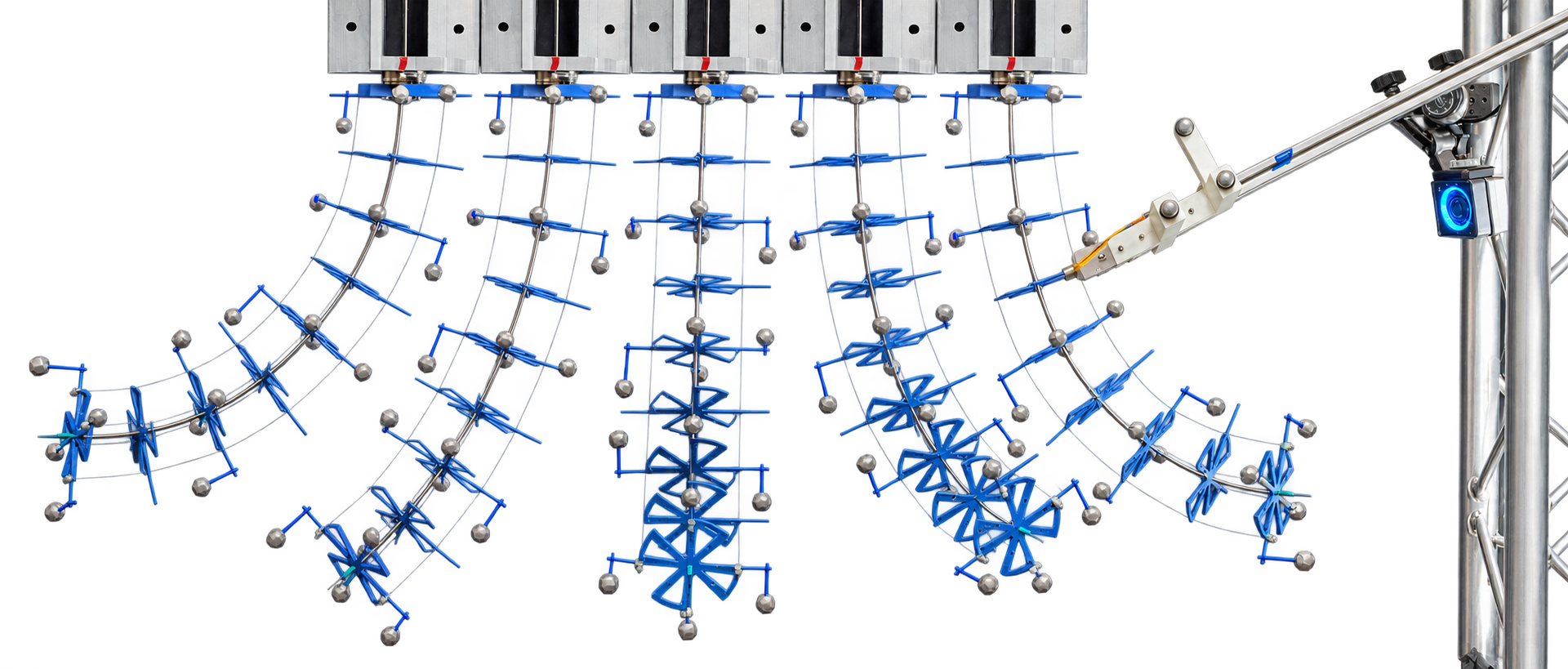}
    };
    
    \node[callout, anchor=east] (A) at (3.6,0.2) {force \\ sensor};
    \draw[callout arrow] (A.west) -- (2.1, 0.2);

    \node[callout, anchor=east] (A) at (-3.0,-1.7) {embedded \\ FBG fiber};
    \draw[callout arrow] (A.east) -- (-2.6, -1.5);
    
    \node[callout] at (2.7,1.6) {contact \\ wand};

    \node[callout] (A) at (-3.8,0.6) {tracked \\ markers};
    \draw[callout arrow] (A.south) -- (-4.1, -0.05);

    \node[callout] (A) at (-3.2,1.55) {tension \\ sensors};
\end{tikzpicture}

%% file: figures/tip_pred_comparison.tikz
\begin{tikzpicture}[
    every node/.style={font=\footnotesize},
    image/.style={
        inner sep=0pt,
        anchor=north west
    },
    label/.style={
        align=center,
        anchor=south,
        font=\footnotesize,
        inner sep=0pt
    }
]

\def\imgheight{0.15\linewidth}
\def\rowsep{-0.5cm}

\node[image] (dynamic) at (0,0) {
    \includegraphics[
        height=\imgheight,
        clip,
        trim=1100pt 400pt 750pt 350pt
    ]{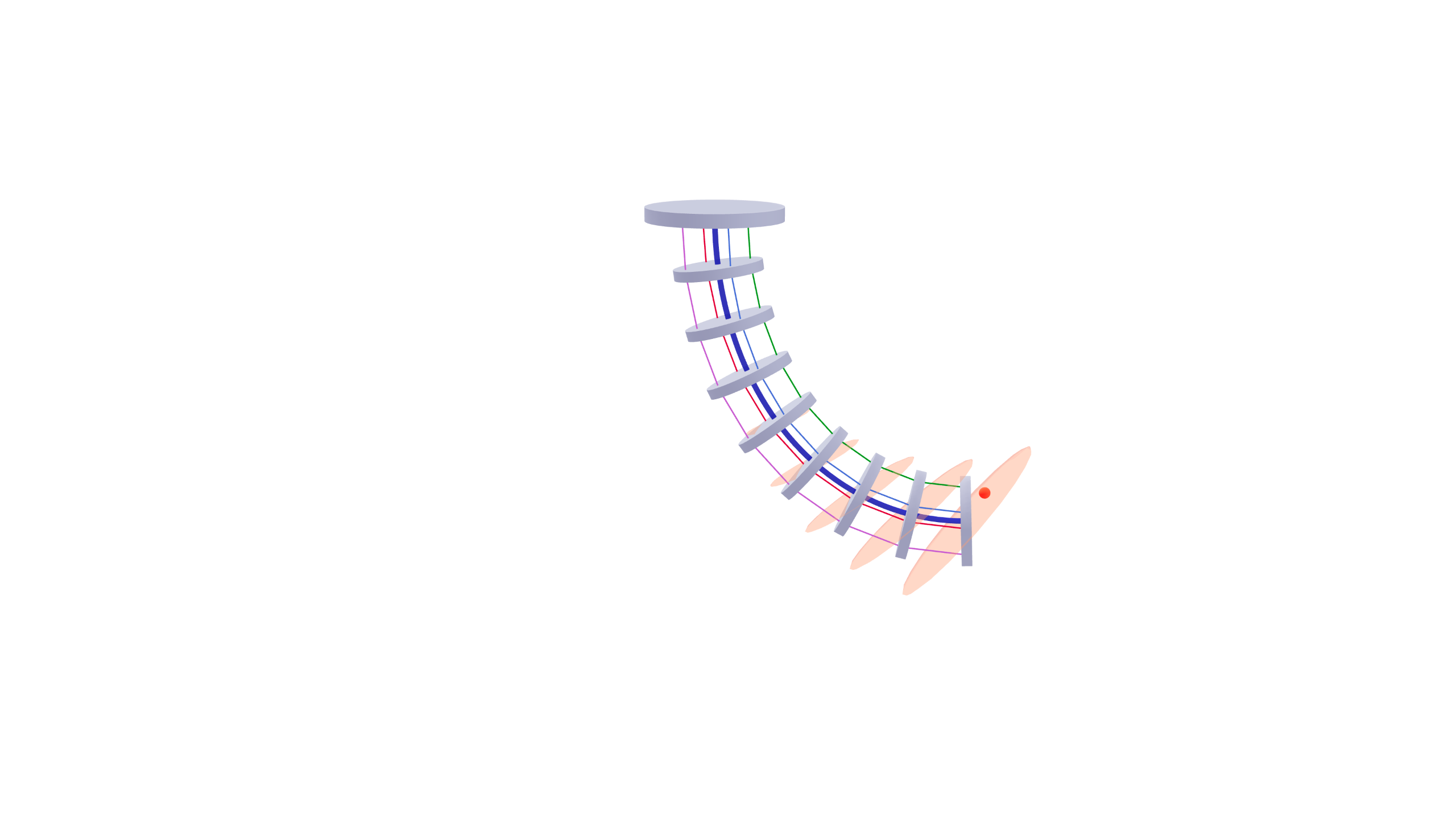}
};

\node[image] (fbg) at ([yshift=\rowsep]dynamic.south west) {
    \includegraphics[
        height=\imgheight,
        clip,
        trim=1100pt 400pt 750pt 350pt
    ]{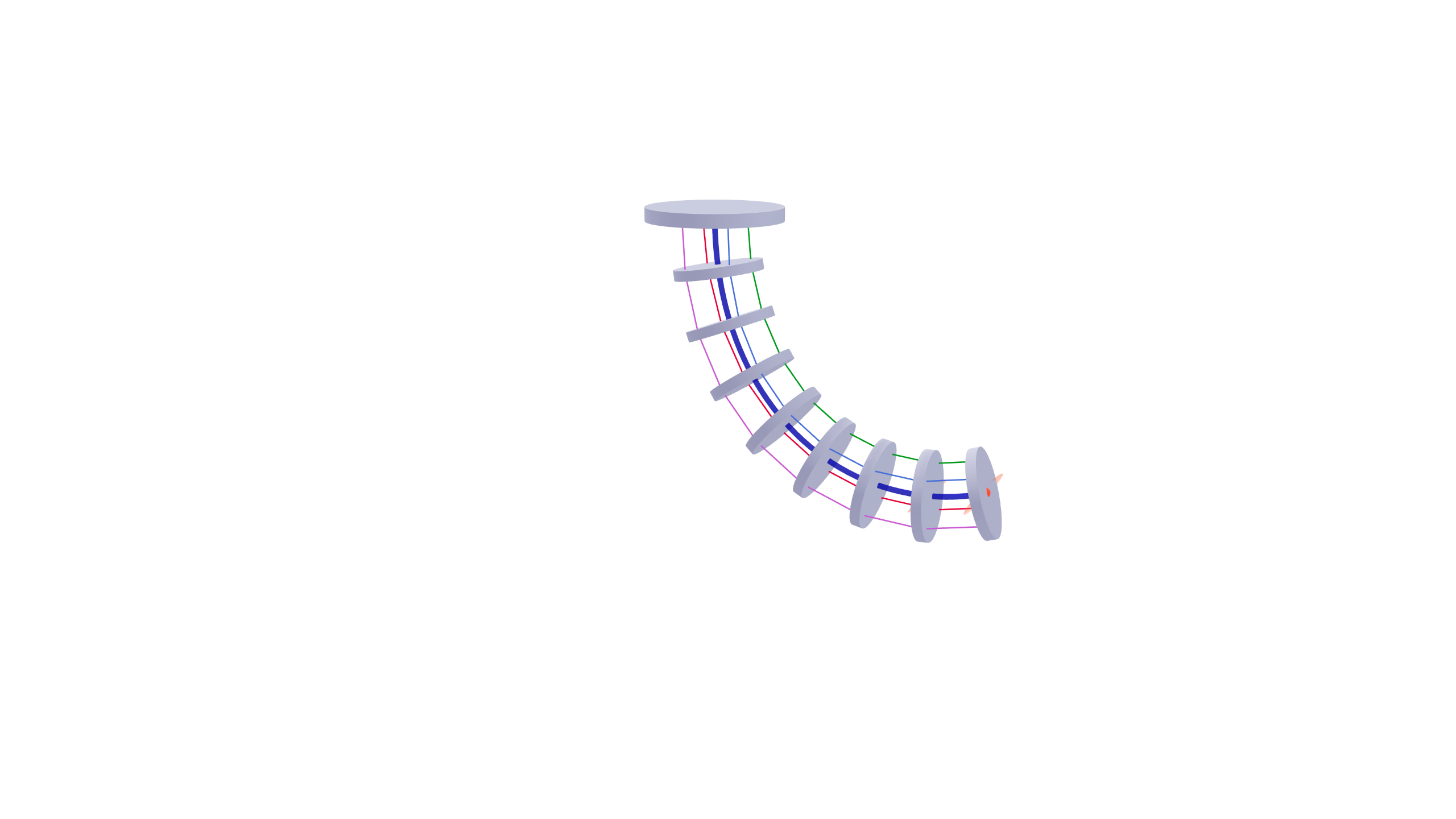}
};

\node[label, xshift=0.6cm, yshift=-0.6cm] at (dynamic.north)
    {Dynamics\\Only};

\node[label, xshift=0.6cm, yshift=-0.6cm] at (fbg.north)
    {Dynamics\\+ FBG};

\end{tikzpicture}